\pdfoutput=1
\documentclass[11pt]{article}

\usepackage[preprint]{acl}

\usepackage{times}
\usepackage{latexsym}
\usepackage[T1]{fontenc}
\usepackage[utf8]{inputenc}
\usepackage{microtype}
\usepackage{inconsolata}
\usepackage{graphicx}
\usepackage{amsmath}
\usepackage{amssymb}
\usepackage{booktabs}

\title{A Deeper Analysis of Block-Sparse Featurizers}

\author{Alexandru-Iulius Jerpelea \\
  Columbia University \\
  \texttt{aij2115@columbia.edu} \And
  Amith Ananthram \\
  Columbia University \\
  \texttt{amith.ananthram@columbia.edu}}

\begin{document}
\maketitle

\begin{abstract}
The recently introduced block-sparse featurizer \citep[BSF;][]{fel2026bsf} is similar to a sparse autoencoder (SAE), but its atomic unit is a small subspace (a block of directions) rather than a single direction. It is designed for features that live on low-dimensional manifolds, which are especially frequent in vision. This work studies the BSF's strengths and weaknesses, finding how it still somewhat suffers from classic SAE failure modes, like feature splitting and composition. We propose several architectural changes to the BSF, including a \emph{Tournament Top-K} selection rule that significantly reduces feature splitting, and we also extend the block paradigm to the crosscoder.
\end{abstract}

\section{Introduction}

When interpreting model activations, it's important to consider the geometry of features. In the literature, the Linear Representation Hypothesis \citep[LRH;][]{park2023linear} posits that models represent features as linear directions in activation space. However, recent work has weakened the picture \citep{engels2024not}, as some features live on low-dimensional manifolds, like days of the week arranged on a circle. By design, SAEs take the LRH for granted, which explains why some of their outputs are confusing (besides other failure modes).

\begin{figure}[t]
  \centering
  \includegraphics[width=\columnwidth]{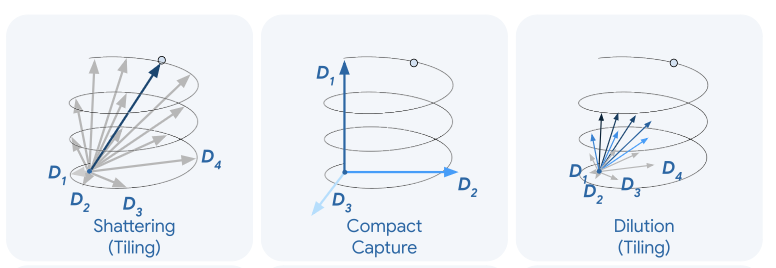}
  \caption{Figure from \citet{bhalla2026manifolds}. Here, the concept manifold is a helix. The first and third pictures show how SAEs capture manifolds. The second picture depicts the ideal capturing regime.}
  \label{fig:tiling}
\end{figure}

More specifically, \citet{bhalla2026manifolds} show that SAEs tile manifolds, spending many directions to cover localities of curved manifolds (Figure~\ref{fig:tiling}). Our own interest in this problem arose in a broader model-diffing study of vision models, where featurizers are needed to compare hidden states; there, we experimented with several featurizers like crosscoders, Matryoshka SAEs, and archetypal SAEs \citep{lindsey2024crosscoders, bussmann2025matryoshka, fel2025archetypal}, and tried to recover concept manifolds post-hoc from our trained SAEs by using atom correlations, Ising-style coupling analysis, etc., but without great success.

\paragraph{Contributions.} This paper focuses on the BSF and the block paradigm it introduces. More concretely, we:
\begin{itemize}
  \item train BSFs on a toy manifold dataset and systematically test for seed stability, spurious correlation resistance, and dominance over classical SAEs;
  \item find that BSFs still ``tile'' manifolds, find a plausible explanation, and propose the \emph{Tournament Top-K} BSF as a response;
  \item propose BSFs with non-constant block dimensions;
  \item present some brief results of us training BSFs on real models;
  \item extend the block paradigm to crosscoders;
  \item discuss how BSFs might capture other types of geometries besides low-dimensional manifolds.
\end{itemize}

\section{The BSF}

Fortunately, the same group recently introduced the block-sparse featurizer \citep[BSF;][]{fel2026bsf}. It's similar to a classic SAE, but its ``code'' is split into small groups of latents (called blocks) and the sparsity penalty is applied to whole blocks rather than to individual directions. This works if we assume a specific underlying hypothesis behind the data-generating process over activations $x \in \mathbb{R}^{d_{model}}$: there is a dictionary of $G$ concepts, where the $g$-th occupies a low-dimensional subspace spanned by an orthonormal frame $D_g \in \mathbb{R}^{b \times d}$ (with $b \ll d$), and each activation is a sparse sum of contributions from a few of them ($S \subseteq \{1,\dots,G\}$, $|S| \ll G$):
\begin{equation}
x = \sum_{g \in S} z_g D_g + \varepsilon,
\end{equation}
where $z_g \in \mathbb{R}^b$ is the activation's position within concept $g$'s manifold and $S$ indexes the concepts present in the input.

We only study the paper's top-$k$ Grassmannian variant, where the encoder is the decoder's transpose (and they are both orthonormal matrices), and the only learned parameters are the frames $D = (D_1, \dots, D_G)$ themselves:
\begin{equation}
z(x) = \Pi_k\big(x D^\top\big), \  \hat{x} = z(x) \cdot D, \  \min_D \lVert x - \hat{x} \rVert_2^2,
\end{equation}
where $\Pi_k$ zeroes all but the $k$ blocks of largest norm $\lVert z_g \rVert_2$. This is basically the SAE literature top-$k$ \citep{gao2024scaling}, but over blocks, not latents. Since $\Pi_k$ sees each block only through its norm, selection is sparse across concepts, but dense within each one, so a whole color circle, for example, can switch on as one atom. This should theoretically reduce manifold tiling.

\section{The Manifold Zoo}

\begin{figure*}[t]
  \centering
  \includegraphics[width=0.85\textwidth]{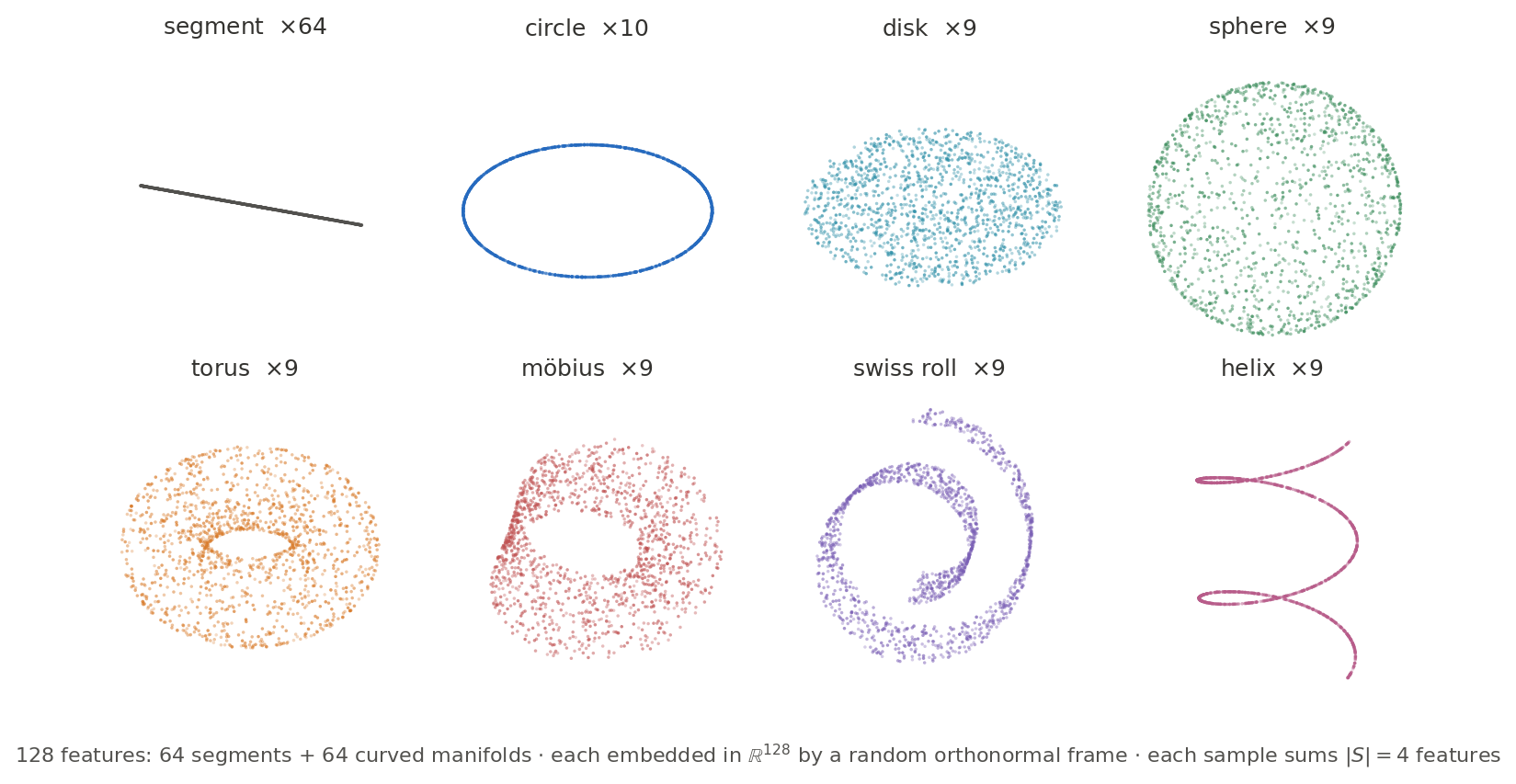}
  \caption{The Manifold Zoo: 128 features (64 segments + 64 curved manifolds), each embedded in $\mathbb{R}^{128}$ by a random orthonormal frame; each sample sums $|S|=4$ features.}
  \label{fig:zoo}
\end{figure*}

Before deploying BSFs on real models, we test them in a setting where we control the data-generating process. The ``Manifold Zoo'' is a synthetic world of $M = 128$ \emph{features}, each with a specific geometry. $64$ of them are straight lines, and the other $64$ are curved manifolds (circles, helices, tori, disks, swiss rolls). Each feature is embedded in a $d$-dimensional activation space ($d = 128$) through a random orthonormal frame. We generate a dataset of $300{,}000$ samples, where every sample is a sparse sum of $|S| = 4$ concepts (Figure~\ref{fig:zoo}).

In order to test the interpretability of a featurizer like the BSF, we match each feature from the dataset to the one block with the highest firing correlation. We can now score how well that block recovers the feature by measuring the $R^2$ between the block's reconstruction and the feature's true contribution to the sample. Sometimes, we also look at how many blocks are needed for a feature to get $R^2 > 0.8$.

Unless stated otherwise, the BSF that we train has $G = 256$ blocks of dimension $b = 3$ with top-$k = 4$ selection. Note how the capacity is generous: with 2 blocks per concept, the sparsity is matched to $|S|$, and dimension $b$ is matched to the largest dimension a manifold can have. In other words, this should be the perfect setting for the BSF to decode each feature's geometry.

We experiment with two variants of the dataset: one where concepts fire independently (\emph{the independent dataset}), and one where we introduce correlation between which concepts fire together (\emph{the correlated dataset}). Feature correlation is a classic pathology for SAE failure, so we consider it necessary to also test this scenario.

\subsection{BSF vs.\ Classic SAE}

First, we want to validate that the BSF truly recovers concept geometry better than an SAE. We train an SAE with an equal number of parameters and similar training settings (top-$k\cdot b$, etc.). To make the comparison fair, we also equip the SAE with post-hoc manifold recovery methods, which, after training, group its atoms into candidate manifolds based on their firing statistics (ValuePCA and partial-correlation grouping). As shown in Table~\ref{tab:vs-sae}, on \emph{the independent dataset} the BSF is superior, but post-hoc recovery works quite decently, too. Once we introduce correlation in the dataset, the BSF is untouched, but the SAE loses accuracy. This is expected, as the post-hoc methods rely on correlation statistics to find the manifolds, so concept co-firing may fool concepts into merging.

\begin{table}[t]
  \centering
  \small
  \begin{tabular}{lcc}
    \toprule
    \textbf{Method} & \textbf{Indep.} & \textbf{Corr.} \\
    \midrule
    SAE (raw atoms) & 0.36 & 0.36 \\
    SAE + post-hoc (ValuePCA) & 0.60 & 0.41 \\
    SAE + post-hoc (partial-corr.) & 0.65 & 0.59 \\
    \textbf{BSF} & \textbf{0.77} & \textbf{0.77} \\
    \bottomrule
  \end{tabular}
  \caption{Mean recovery $R^2$ of the BSF against a parameter-matched SAE, with and without post-hoc manifold recovery, on the independent and correlated datasets.}
  \label{tab:vs-sae}
\end{table}

\subsection{BSF Consistency Across Seeds}

A big problem with SAEs is that they are not stable, as identical models trained with different seeds can produce significantly different dictionaries (see archetypal SAEs; \citealp{fel2025archetypal}). We test BSFs by training 10 of them on the same data, varying only the seed.

\begin{figure*}[t]
  \centering
  \includegraphics[width=\textwidth]{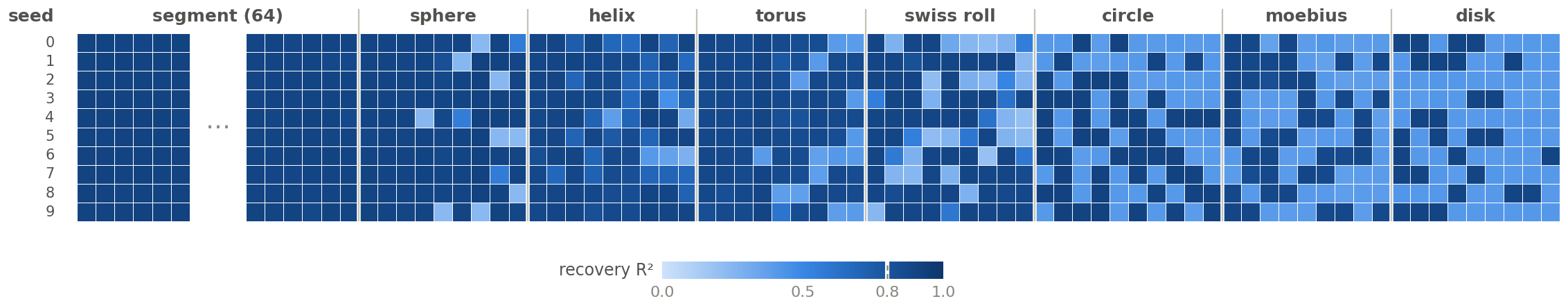}
  \caption{Per-seed recovery $R^2$ of each feature (columns) across 10 seeds (rows): segments, spheres, helices, and tori are almost always well recovered, while the other manifolds flicker from seed to seed.}
  \label{fig:seeds}
\end{figure*}

In Figure~\ref{fig:seeds}, we match each feature to the block whose firing correlates most with it, and measure how well that block actually recovers the feature. We observe that segments, spheres, helices, and tori are almost always well recovered by one block, while the other concept manifolds flicker from seed to seed.

\begin{figure*}[t]
  \centering
  \includegraphics[width=\textwidth]{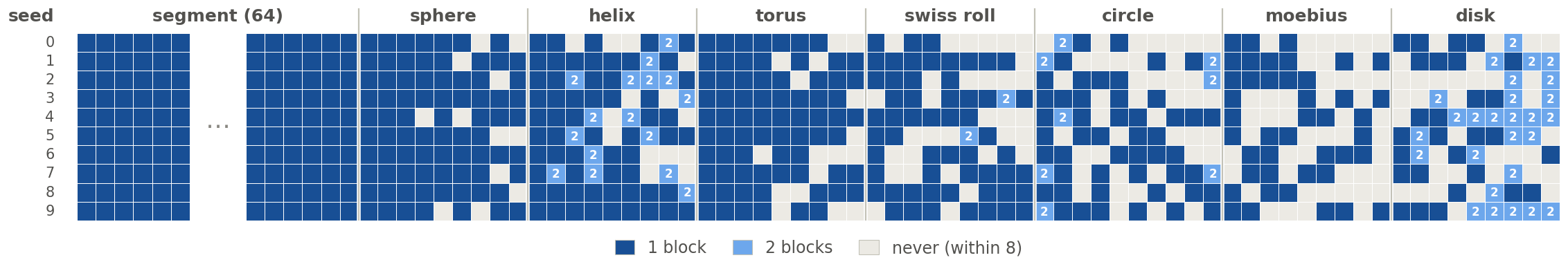}
  \caption{Number of blocks each feature needs to reach $R^2 > 0.8$, per seed: either 1, 2, or never (within 8).}
  \label{fig:blocksneeded}
\end{figure*}

We now ask how many blocks a feature needs to reach $R^2 > 0.8$. The answer is either 1, 2, or more than 8 (which we count as not recovered); see Figure~\ref{fig:blocksneeded}. Interestingly, every concept is captured by exactly one block in at least one of the 10 seeds, suggesting that no feature is intrinsically impossible to recover.

\begin{figure}[t]
  \centering
  \includegraphics[width=0.85\columnwidth]{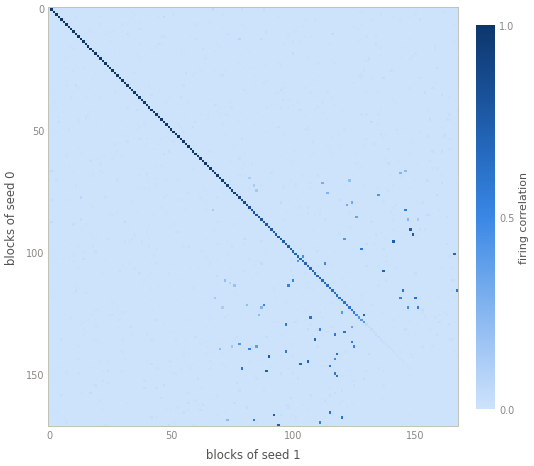}
  \caption{Firing correlation between the blocks of two seeds: close to an identity, fading for blocks whose features tend to split.}
  \label{fig:seedcorr}
\end{figure}

Finally, we pick two seeds and compute the correlation between their blocks' firings. The correlation matrix is quite clean, close to an identity. The identity fades for those blocks whose corresponding features are more likely to split across multiple blocks (Figure~\ref{fig:seedcorr}).

\begin{figure*}[t]
  \centering
  \includegraphics[width=\textwidth]{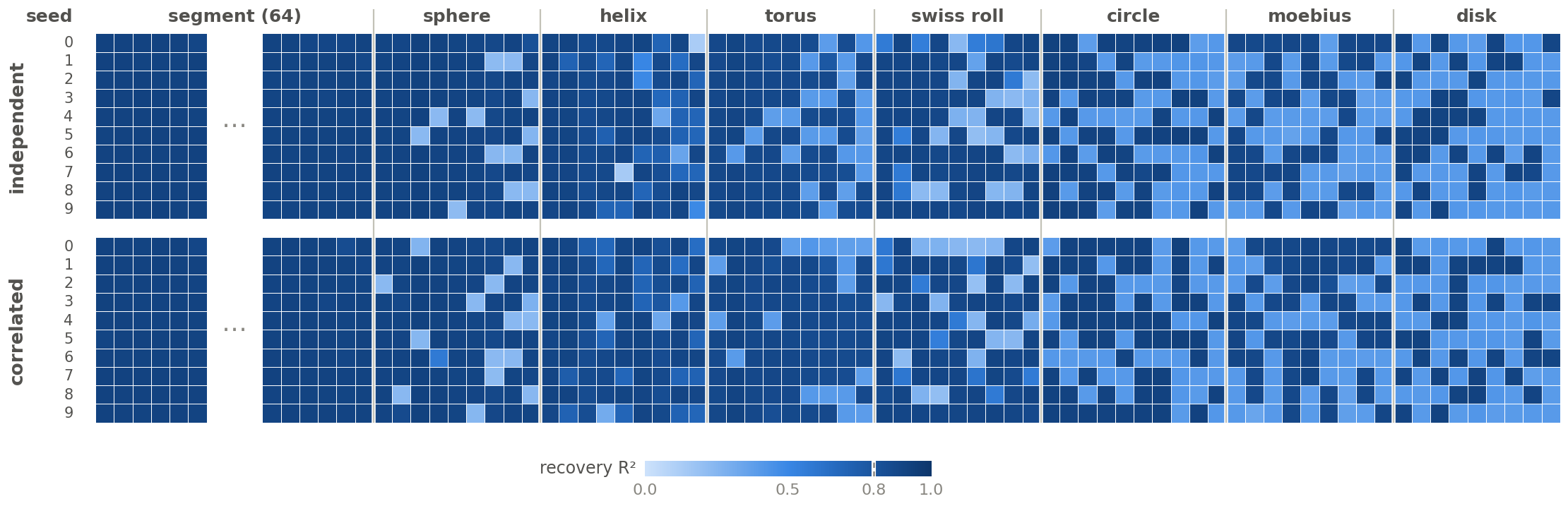}
  \caption{The 10-seed experiment repeated on the correlated dataset: similar $R^2$ scores and similar feature splitting patterns as on the independent dataset.}
  \label{fig:seedscorr}
\end{figure*}

We also repeat the identical 10-seed experiment on \emph{the correlated dataset}. The results are basically the same, with similar $R^2$ scores, and similar feature splitting patterns (Figure~\ref{fig:seedscorr}).

\subsection{Testing if Hyperparameters Cause Splitting}

So we have shown that BSFs are pretty consistent, but while doing so, we discovered that some features arbitrarily split among multiple blocks, even on \emph{the independent dataset}, where there is no spurious correlation between features. This is quite reminiscent of how classical SAE directions tile manifolds. In our case, manifolds split among blocks.

\begin{figure*}[t]
  \centering
  \includegraphics[width=\textwidth]{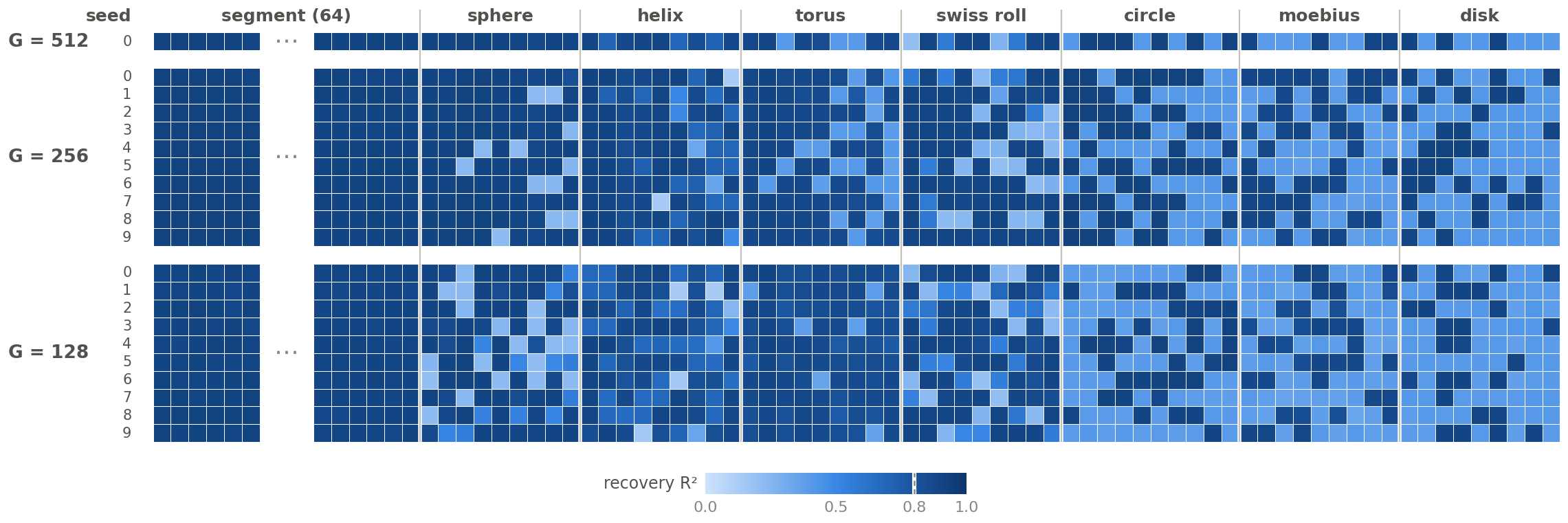}
  \caption{Sweeping the number of blocks $G \in \{128, 256, 512\}$: splitting persists in every setting, even at $G = M = 128$.}
  \label{fig:gsweep}
\end{figure*}

Two big factors of tension in any SAE are its capacity (total number of atoms) and its sparsity, and both are known to shape its pathologies \citep{chanin2024absorption, gao2024scaling}. Thus, we wonder if our hyperparameters cause splitting, so we first sweep $G \in \{128, 256, 512\}$, but to no avail (Figure~\ref{fig:gsweep}). Not even $G = M = 128$ works, i.e., having as many blocks as features, where we would have hoped that we could get a feature-block bijection.

\begin{figure*}[t]
  \centering
  \includegraphics[width=\textwidth]{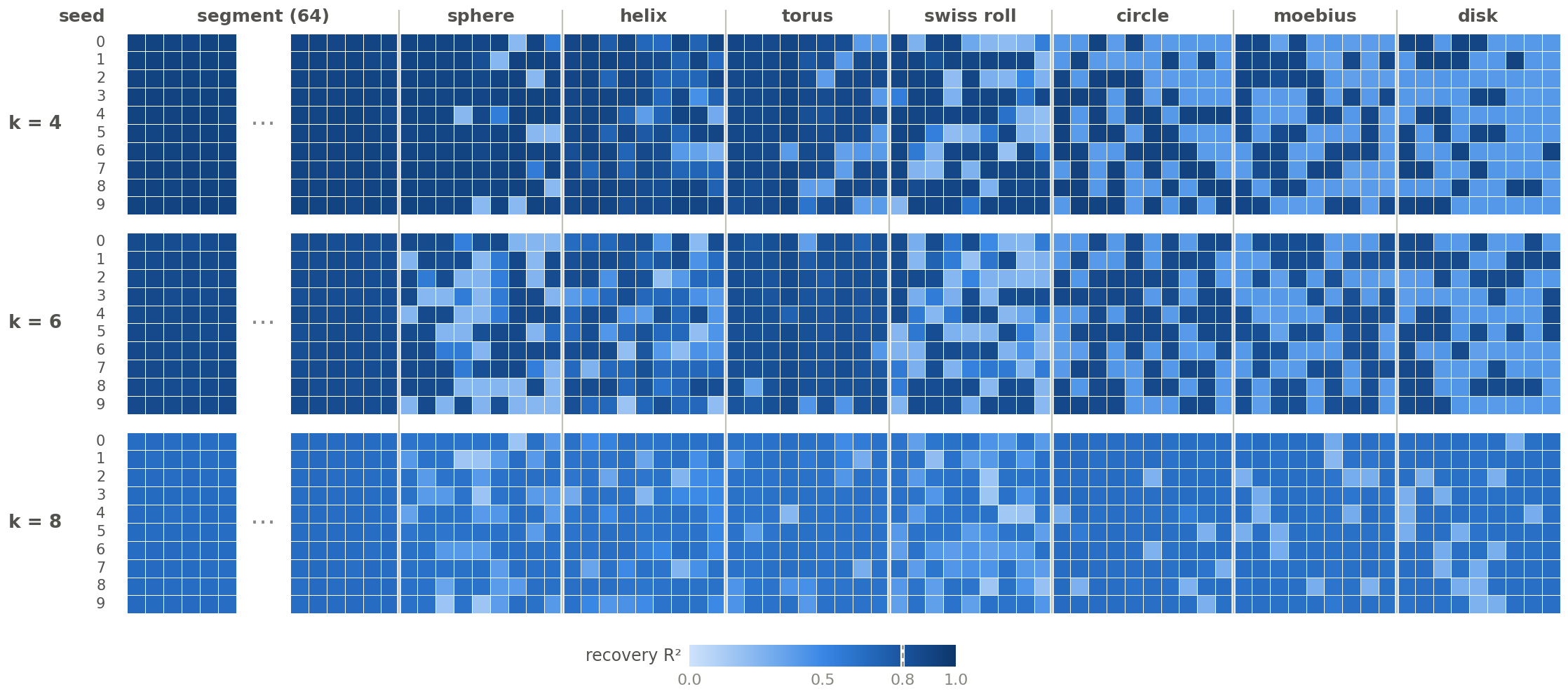}
  \caption{Sweeping sparsity $k \in \{4, 6, 8\}$: performance degrades as $k$ increases; at $k=8$ every feature splits across multiple blocks.}
  \label{fig:ksweep}
\end{figure*}

We also vary sparsity. As our featurizer uses top-$k$, we sweep $k \in \{4, 6, 8\}$, but that only degrades performance as $k$ increases, which is expected as $|S| = 4$. For $k = 8$, not a single feature is captured by one block anymore, as they all split across multiple blocks (Figure~\ref{fig:ksweep}).

\subsection{So Why Do Splits Form?}

\begin{figure*}[t]
  \centering
  \includegraphics[width=0.85\textwidth]{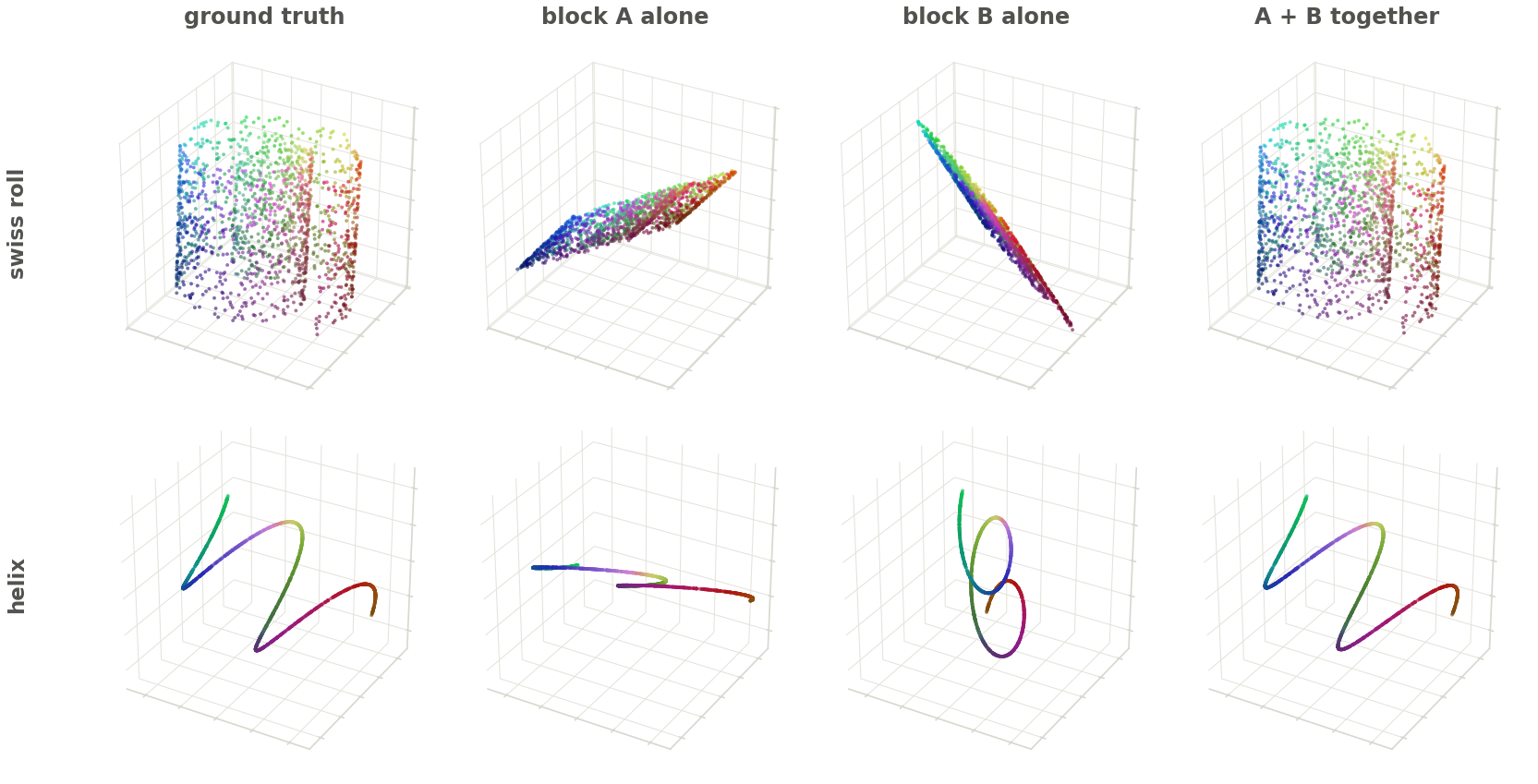}
  \caption{Visualizing split features (one swiss roll, one helix): ground truth, what each of the two blocks captures alone, and their sum. The two blocks tend to be near-orthogonal complements.}
  \label{fig:splitviz}
\end{figure*}

Let's try and visualize some of the features that split across two blocks. In Figure~\ref{fig:splitviz}, every dot is a sample where the feature is active (we look at one swiss-roll and one helix feature), colored by its position on the manifold. A dot keeps its color in every panel of the row. The two middle columns show what each of the two blocks capture of the feature, and the last column is their sum. We observe that when a manifold splits among two blocks, the blocks tend to be near-orthogonal complements.

\begin{figure}[t]
  \centering
  \includegraphics[width=\columnwidth]{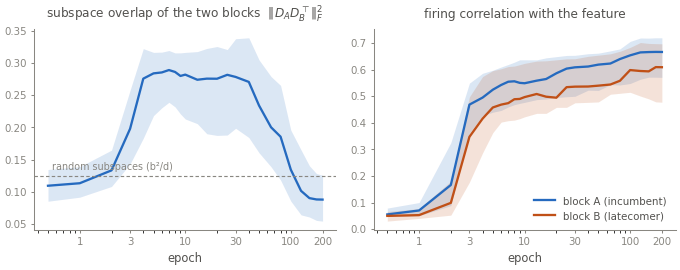}
  \caption{Tracking 83 split block pairs through training: subspace overlap $\lVert D_A D_B^\top \rVert_F^2$ (left) and each block's firing correlation with the feature (right).}
  \label{fig:pairtracking}
\end{figure}

As previously noted, when a feature splits along multiple blocks, it's almost always exactly 2. For 83 such block pairs, we track the pair through BSF training. The first panel of Figure~\ref{fig:pairtracking} shows the geometric overlap between the two blocks' subspaces, and the second shows each block's firing correlation with the feature (blue for the first block locked onto the feature, and red for the second block).

First, see how both blocks start specializing on the same feature around the same time ($\sim$epoch 3). For a while, their subspaces grow similar, at $\sim$0.3 overlap (2x of the random baseline, so definitely not copies of one another). But then they pull apart, ending up to be almost orthogonal components of the same feature.

We guess that in the first epochs blocks are learning the feature's strongest axes of variation, and that's why they grow together. After that sharp growth, both blocks have high correlation with the feature, so they are officially ``locked'' to it, but they are still significantly different. Each block gets gradient from the samples it fires on, so each gets reinforced exactly on the differences it has compared to the other, which drives the blocks to become almost orthogonal. Once they split, each covers variation in samples that the other misses, so gradient descent has no incentive to eliminate one of them.

But we don't fully understand this phenomenon. Given that the pair blocks decompose the manifold orthogonally rather than discovering different regions, they would best reconstruct a sample if they fired together, yet $k = |S|$ discourages multiple blocks per feature at once, so the pair of blocks tends to not fire together that often (although they do fire together above chance). We also ablate \emph{Aux-K} (an auxiliary loss standard for top-K SAEs; \citealp{gao2024scaling}) and \emph{k-annealing}, and find that none of them significantly contributes to block splitting.

So, just as SAEs tile manifolds, BSFs also split manifolds (although at a smaller rate). We are unsure of the reason, but discover the useful pattern of how block pairs evolve, which we will use in the following section to develop a new BSF architecture.

\section{A Solution to Feature Splitting: Tournament Top-K}

A lot of SAE problems like feature composition, hedging, absorption, etc.\ imply that some SAE latents share directions (i.e., they are not orthogonal). In our case, feature splitting causes the same thing, so it is reasonable to look into known SAE solutions. For example, MP-SAEs \citep{costa2025matching} decode by greedily subtracting atoms, but this does encourage a new way of feature composition. Ort-SAEs \citep{korznikov2025ortsae} penalize similarities between all pairs of latents, but this is expensive and we also found it to be too discriminatory in the first stages of training. However, we are inspired by Ort-SAEs to propose \emph{Tournament Top-K}:

\begin{quote}
\emph{For each sample, go through the blocks in decreasing order of magnitude $|z_g|$, keeping a running set $B$ of accepted blocks. For the next block $g$, if some $h \in B$ overlaps with its geometry sufficiently (we say $\Omega_{gh} > 0.1$), then we skip $g$ and count a win for $h$. Otherwise, add $g$ to $B$. We stop at $|B| = k$ (the BSF's sparsity).}

\emph{Note that blocks $g$ and $h$ could meet again in a future duel. However, if at some point $\Omega_{gh} > 0.5$, the block with more wins will permanently win the feature, which means that in any other future duel we will keep the winning block in $B$, and exclude the other.}
\end{quote}

\begin{figure}[t]
  \centering
  \includegraphics[width=0.9\columnwidth]{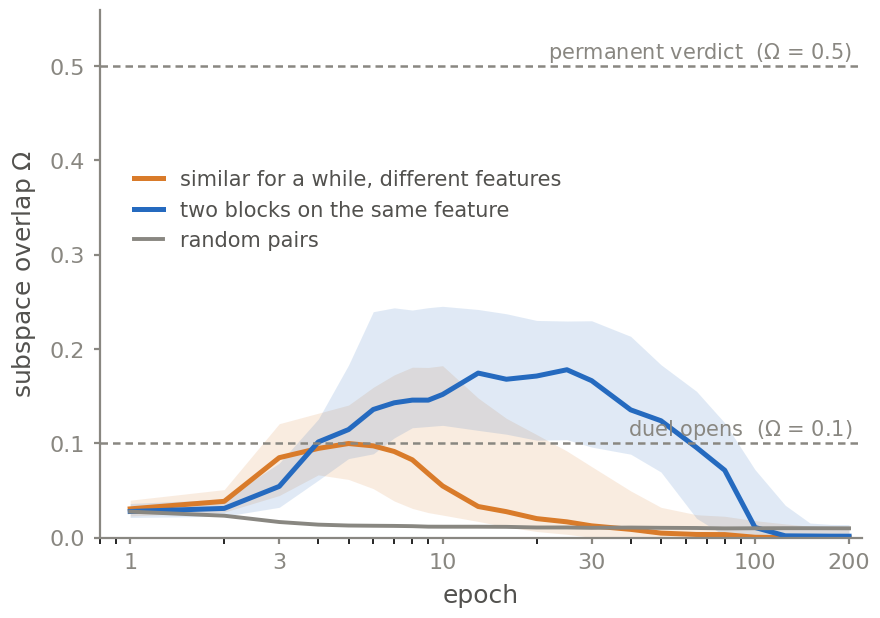}
  \caption{Subspace overlap $\Omega$ through training for block pairs that are similar for a while but settle on different features (orange), pairs that actually split one feature (blue), and random pairs (gray), with the duel-opening ($\Omega = 0.1$) and permanent-verdict ($\Omega = 0.5$) thresholds.}
  \label{fig:duel}
\end{figure}

To understand why we are using this 2-threshold system, consider Figure~\ref{fig:duel} (note that $\Omega$ is a different metric for geometric similarity than the one used above, comparing only directions blocks actually use, thus we have a different scale).

In the first epochs, there are many pairs of blocks that overlap, yet they naturally set on different features later (orange line). However, they are initially indistinguishable from blocks that actually split features (blue line). Thus, it's hard to set just one threshold. Our initial $0.1$ threshold catches true splits about just a third of the time. That's why we make it reversible (a winner is not decided yet). Also for this reason, we only begin dueling after a warmup of $\sim$16 epochs, allowing blocks to specialize a bit.

Note that the $0.5$ threshold is naturally uncrossable by pairs of blocks. However, our $0.1$ duels push blocks up there: if two blocks really share a feature, they get chosen in an alternate manner by the $0.1$ duel (depending on the sample), which pushes them both to reconstruct the whole feature alone. The overlap thus climbs past $0.5$; if two blocks don't actually share a feature, the $0.1$ duel exclusivity doesn't actually do anything, so each block just develops on its own, and they never meet again in a duel.

\begin{figure*}[t]
  \centering
  \includegraphics[width=\textwidth]{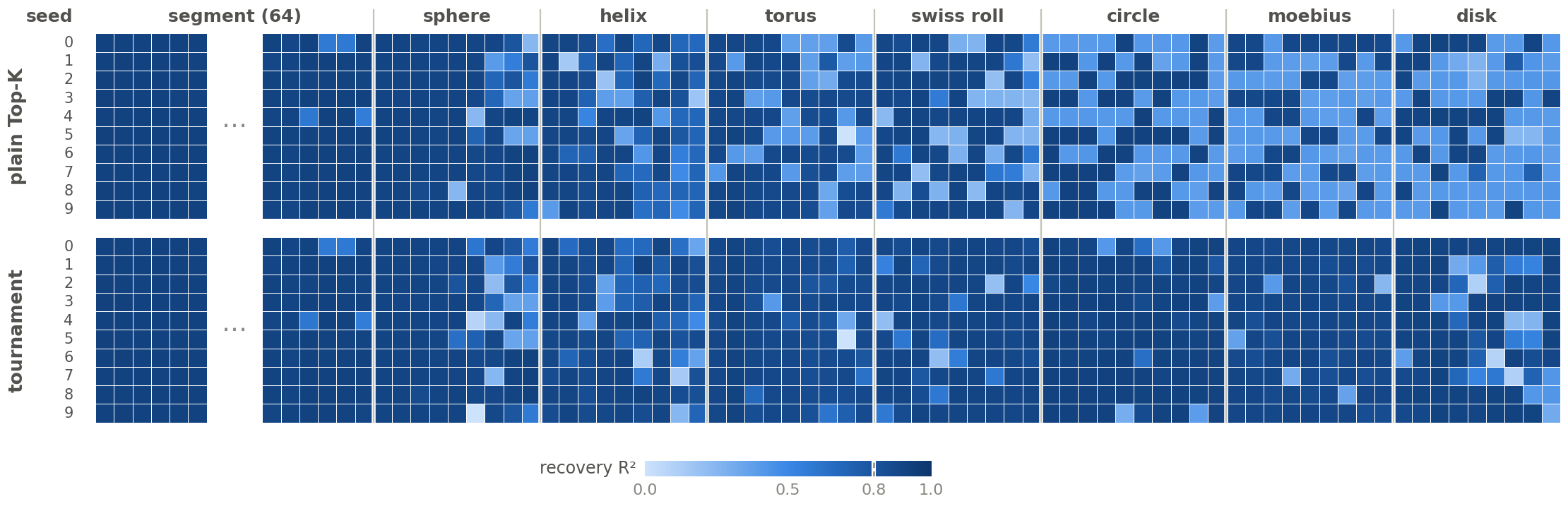}
  \caption{Plain top-K vs.\ Tournament Top-K on the correlated dataset: recovery $R^2$ per feature and seed. Tournament Top-K lifts the fraction of well-recovered manifolds from $0.77$ to $0.90$.}
  \label{fig:tournament}
\end{figure*}

We test on \emph{the correlated dataset}, which is our hardest setting. Tournament Top-K raises mean recovery $R^2$ significantly, and also lifts the fraction of well-recovered manifolds from $0.77$ to $0.90$, significantly cutting the splitting phenomenon (Figure~\ref{fig:tournament}).

\section{Varied-Dimension Blocks}

We also note that feature manifolds have varied dimensionalities, yet all blocks are 3D. This leads to feature merging: for example, we observed cases of up to three linear features being explained by one block. We thus give our BSF a mix of block sizes (128 $\times$ 1D + 38 $\times$ 2D + 90 $\times$ 3D blocks), and test whether training self-sorts. We try the following regimes (and fail in each of them):

\begin{itemize}
  \item \emph{vanilla top-K}: 3D blocks have higher norms on average, so they win the top-K races often. Segments mostly end up in 3D blocks, and curved features collapse.
  \item \emph{normalized top-K}: divides each block magnitude by its expected scale $\sqrt{b_g}$, with $b_g \in \{1, 2, 3\}$, before top-K selection. Produces the reverse phenomenon, where 1D blocks are disproportionately preferred.
  \item \emph{MDL score}: subtracts $\lambda \cdot b_g$ from each block's magnitude before top-K selection, with tunable $\lambda$. Fails similarly.
  \item \emph{JumpReLU with L0 penalty}: we thought that maybe top-K is incompatible with our loss terms, so we tried another architecture, but still failed.
\end{itemize}

We measured that a BSF with a mix of block sizes would reach a smaller loss than a regular BSF if it learned how to map features to blocks optimally, yet it seems training can't reach this. Once a feature sticks to a block in the first epochs, moving it is impossible for gradient descent, because a block move raises the loss for a significant time. We think interventions in the spirit of Tournament Top-K might be needed, \emph{but we leave that as a homework for the reader.}

\section{Applying BSFs on Real Models}

We have also trained BSFs on real models (DINOv3 and CLIP; \citealp{simeoni2025dinov3, radford2021clip}). We swept through many configurations to conclude our best setting: $16{,}384$-dimensional dictionaries, with top-$K = 64$, and blocks of dimension $3$. We also train Matryoshka SAEs \citep{bussmann2025matryoshka} at the same scale. Some of the experiments we run with the BSFs:

\begin{itemize}
  \item \emph{cross-model matching}: pairing blocks between CLIP and DINO by activation correlation. We get a few hundred strongly matched blocks, and found that truly private blocks are rare.
  \item \emph{artificial features}: we render synthetic images varying color, position, orientation, frequency, and shape, and found dedicated blocks carrying each of them.
  \item \emph{causality}: we ablate and inject blocks on labeled external data (ImageNet). Blocks are causal to their features, but we figured out that most features are split among multiple blocks, so we had to ablate across quite a few blocks at once to observe the causality on a preset of features.
  \item \emph{layer-to-layer maps}: we train linear maps between layers and use BSFs to read what these maps preserve.
\end{itemize}

Note that this section is very brief. The point of this paper is to study the BSF itself, while the aforementioned experiments diverge from that. Many of these will be further detailed in our follow-up model diffing paper.

\section{Block Crosscoders}

Crosscoders \citep{lindsey2024crosscoders} are a really important mechanistic interpretability tool for model diffing, and we wonder if the block paradigm extends to them. In this experiment, we maintain the top-K method. Consider a paired sample $(x^A, x^B)$ of hidden states from some layers in models $A$ and $B$. We get the code by doing:
\begin{equation}
z = \Pi_k\left(x^A E^A + x^B E^B\right),
\end{equation}
where $\Pi_k$ keeps the $k$ blocks of largest norm $\lVert z_g \rVert_2$, exactly as before. The code then decodes through two dictionaries,
\begin{equation}
\hat{x}^m = \sum_{g \in \Pi_k} z_g D_g^m, \qquad m \in \{A, B\},
\end{equation}
with the objective $\min_{D} \sum_m \lVert x^m - \hat{x}^m \rVert_2^2$, where $\Pi_k$ selects the $k$ blocks of largest joint norm.

\subsection{Merging Problem}

We do notice a structural problem with this simple extension of the crosscoder though. Testing it on a new Manifold Zoo, with 64 shared features, 32 exclusive to $A$, and 32 exclusive to $B$, we find that the crosscoder often assigns an $A$-only feature to block $g$, while also assigning a $B$-only feature to the same block $g$.

Note how this problem is specific to the block-crosscoder. In a normal crosscoder, a latent (a dimension of $z$) is a single scalar, so both decoders multiply by the same scalar. Thus, the two concepts would have to fire together with the same intensity at all times. In a block however, the corresponding $z_g$ has $b$ dimensions, so one decoder can use some of them, while the other decoder uses the rest. This is especially lucrative when the sum of the dimensionalities of the $A$-only and $B$-only features is $\le b$. That way, the two concepts can stay fully independent within the same block.

\subsection{Dedicated Feature Columns}

To fix this, we use dedicated feature columns \citep[DFCs;][]{jiralerspong2026crossarch} from the crosscoder literature. This method pre-partitions blocks into 3 pools: shared, $A$-only, and $B$-only. So basically $D^A$ is zeroed-out on the $B$-only blocks, and vice versa. On our toy dataset, feature collisions become much rarer with block-DFC crosscoders.

\subsection{Results}

We train a block-DFC crosscoder on paired DINO/CLIP activations at layer 10, with $G = 8192$ blocks, and a pool distribution of 4096 shared blocks, 2048 DINO-only, and 2048 CLIP-only. EVs are healthy at 0.71 for DINO and 0.66 for CLIP, and we only have 1 dead block.

We now match each single-model BSF's blocks with the crosscoder's blocks, by firing rate correlation (Table~\ref{tab:crosscoder}).

\begin{table}[t]
  \centering
  \small
  \begin{tabular}{lcc}
    \toprule
     & \textbf{DINO BSF} & \textbf{CLIP BSF} \\
    \midrule
    Blocks recovered & \textbf{82\%} & 53\% \\
    Pool (sh.\ / DINO / CLIP) & 26 / 73 / 1 & 62 / 5 / 33 \\
    \bottomrule
  \end{tabular}
  \caption{Matching each single-model BSF's blocks to the block-DFC crosscoder's blocks by firing rate correlation: fraction of blocks recovered, and which crosscoder pool (shared / DINO-only / CLIP-only, in \%) the recovered blocks come from.}
  \label{tab:crosscoder}
\end{table}

It looks like the crosscoder really learns DINO's dictionary and leverages its exclusive pool, but treats CLIP as mostly shared and does not even cover it well. It's an open question whether CLIP's visual features are a subset of DINO's, or CLIP's code is just harder to recover.

\section{Conclusion}

The BSF delivers on its core promise: when features live on manifolds, it recovers them as single units, while regular SAEs tile them into fragments. Still, it does not escape the classic SAE pathologies, as features keep splitting across blocks even with no correlations in the data. The splitting has structure though (split blocks end up near-orthogonal complements of each other), and our Tournament Top-K exploits this geometry to recover much of what vanilla top-k loses. Mixing block dimensionalities fails, with features locking into wrong-sized blocks early in training. Extending blocks to crosscoders brings a merging problem, which dedicated feature columns partly fix. Overall, we think the BSF is a step towards the right direction, i.e., developing featurizers suited to the data generating process. We are going to use the BSF intensely in our follow-up model diffing paper.

\bibliography{custom}

\end{document}